\documentclass[letterpaper]{article} 
\usepackage{aaai24}  
\usepackage{times}  
\usepackage{helvet}  
\usepackage{courier}  
\usepackage[hyphens]{url}  
\usepackage{graphicx} 
\usepackage{natbib}  
\usepackage{caption} 
\usepackage{algorithm}
\usepackage{algorithmic}

\usepackage{newfloat}
\usepackage{listings}
\DeclareCaptionStyle{ruled}{labelfont=normalfont,labelsep=colon,strut=off} 
\floatstyle{ruled}
\newfloat{listing}{tb}{lst}{}
\floatname{listing}{Listing}
\title{Decoupling Representation and Knowledge for Few-Shot Intent Classification and Slot Filling}
\author{
    Jie Han\textsuperscript{\rm 1},
    Yixiong Zou\textsuperscript{\rm 1}\thanks{Corresponding author.},
    Haozhao Wang\textsuperscript{\rm 1}, 
    Jun Wang\textsuperscript{\rm 2}, 
    Wei Liu\textsuperscript{\rm 1}, 
    Yao Wu\textsuperscript{\rm 3}, 
    Tao Zhang\textsuperscript{\rm 3}, 
    Ruixuan Li\textsuperscript{\rm 1}\footnotemark[1]
}
\affiliations{
    \textsuperscript{\rm 1}School of Computer Science and Technology, Huazhong University of Science and Technology\\
    \textsuperscript{\rm 2}iWudao Tech\\
    \textsuperscript{\rm 3}Banma Network Technology\\

    \{jiehan,yixiongz,hz\_wang,idc\_lw,rxli\}@hust.edu.cn, jwang@iwudao.tech, \{qifang.wy,billow.zhangt\}@alibaba-inc.com\\
}

\usepackage{bibentry}

\usepackage{color}
\usepackage{amssymb}
\usepackage{amsmath}
\usepackage{booktabs}
\usepackage{multirow}

\begin{document}

\maketitle

\begin{abstract}
Few-shot intent classification and slot filling are important but challenging tasks due to the scarcity of finely labeled data. Therefore, current works first train a model on source domains with sufficiently labeled data, and then transfer the model to target domains where only rarely labeled data is available.
However, experience transferring as a whole usually suffers from gaps that exist among source domains and target domains. For instance, transferring domain-specific-knowledge-related experience is difficult.
To tackle this problem, we propose a new method that explicitly decouples the transferring of general-semantic-representation-related experience and the domain-specific-knowledge-related experience.
Specifically, for domain-specific-knowledge-related experience, we design two modules to capture intent-slot relation and slot-slot relation respectively.
Extensive experiments on Snips and FewJoint datasets show that our method achieves state-of-the-art performance. The method improves the joint accuracy metric from 27.72\% to 42.20\% in the 1-shot setting, and from 46.54\% to 60.79\% in the 5-shot setting.
\end{abstract}

\section{Introduction}
Natural language understanding (NLU) is a critical component of conversational dialogue systems, such as Siri, Alexa, and Google Assistant. Specifically, NLU includes two sub-tasks: (1) intent classification, which classifies an utterance into an intent label, and (2) slot filling, which classifies each word in the utterance into a slot label. 
Both sub-tasks rely on large amounts of finely labeled data, which is difficult to obtain. 
Therefore, some work proposed to study NLU in the setting of few-shot, where only a few labeled data are available~\cite{krone-etal-2020-learning, gangadharaiah2022zeroshot}. 
Recently, to boost NLU learning in the few-shot setting, some methods tried to utilize labeled data from other sufficiently labeled source domains to help NLU learning (in the rarely labeled target domains)~\cite{rongali-etal-2023-low, kwon2023zeroshot}. 
However, due to gaps that exist among different domains, it is hard to directly transfer prior experience from source domains to target domains.

In this paper, we hold that the prior experience in source domains consists of two parts: the \textbf{general semantic representation} that refers to utterance semantics, and the \textbf{domain-specific knowledge} that refers to the intent-slot relation and slot-slot relation. Intuitively, general semantics are shared across domains while the domain-specific intents/slots are not. Therefore, it is reasonable to transfer the general semantic representation from source domains to target domains, while it is difficult to transfer the domain-specific knowledge. 
Despite a few work has noticed the differences between the two parts, they still tried to conduct the transferring as a whole~\cite{hou-etal-2021-learning,liu-etal-2021-explicit-joint}, which may be inefficient.

\begin{figure}[t]
\centering
\includegraphics[width=1.0\columnwidth]{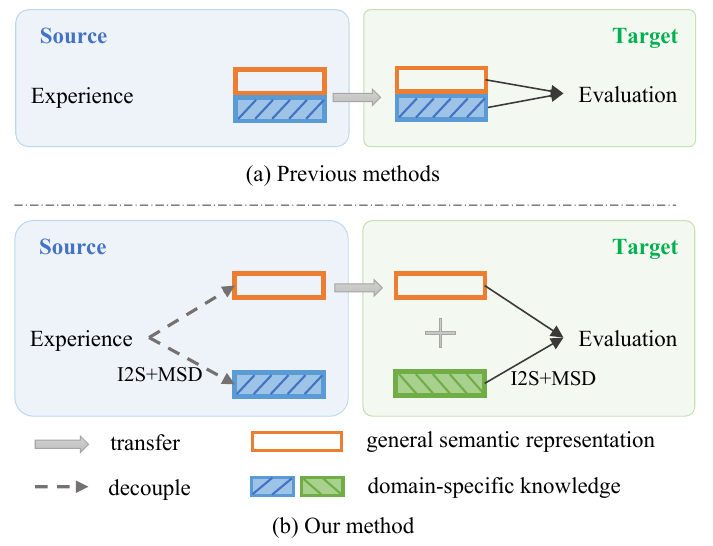}
\caption{Framework of Joint Modeling with Relationship Masking (JMRM).
Firstly, we explicitly decouple the general semantic representation and the domain-specific knowledge of source domains. 
Secondly, we only transfer the general semantic representation to target domains. Thirdly, we apply the knowledge specific to target domains to the transferred representations.
}
\label{fig_introduction}
\end{figure}

To this end, we propose \textbf{Joint Modeling with Relationship Masking} (\textbf{JMRM}), which explicitly decouples the transferring of the general semantic representation and the domain-specific knowledge. 
Specifically, to achieve decoupling, we design two modules, \textbf{I2S-Mask} (\textbf{I2S}) and \textbf{Masked Slot Decoding} (\textbf{MSD}), to capture the domain-specific knowledge (intent-slot relation and slot-slot relation). 
In the I2S module, we introduce an intent-slot correlation score matrix to capture the intent-slot relation. Here,
the matrix is automatically summarized from the labeled data. Furthermore, it regularizes the predicted intent and slot labels that are related.
For example, as shown in Figure~\ref{fig_method}, in the predictions of an utterance, each intent label can only co-occur with the slot labels related to this intent, such as $PlayMusic$ and $B\mbox{-}artist$, and an intent label should not co-occur with the slot labels unrelated to the intent, such as $PlayMusic$ and $B\mbox{-}city$.
In the MSD module, we introduce a slot-slot constraint score matrix to capture the slot-slot relation. Here, the matrix is automatically summarized from BIO annotation rules. Furthermore, it constrains the predicted slot label sequence is rational.
For example, as shown in Figure~\ref{fig_method}, in the predicted slot label sequence, $I\mbox{-}artist$ is allowed to follow $B\mbox{-}artist$, but not $O$.

Specifically, we implement the above design as two masking operations applied during both training and testing. During source-domain training, such operations will decouple the general semantic representation and the domain-specific knowledge. During target-domain testing, such operations will efficiently summarize target-domain information, and apply them to the transferred general representations, which therefore helps the target-domain recognition.

Experiments on two public datasets, the Snips dataset~\cite{Snips} and the FewJoint dataset~\cite{FewJoint}, show that JMRM achieves state-of-the-art performance. 
Specifically, JMRM improves the joint accuracy from 27.72\% to 42.20\% in the 1-shot setting, and from 46.54\% to 60.79\% in the 5-shot setting.
Furthermore, since JMRM is a plug-and-play method, we demonstrate that it can improve the effectiveness of other models.
Moreover, extensive analysis suggests that decoupling the transfer of the general semantic representation and the domain-specific knowledge is more efficient than transferring as a whole.

In summary, the main contributions of this paper are summarized as follows:
\begin{enumerate}
    \item To relieve the difficulty of transferring caused by the gaps between different domains, we try to explicitly decouple the general semantic representation and the domain-specific knowledge.
    \item We propose the JMRM method, which contains I2S and MSD modules. The two modules explicitly utilize two relationship score matrices to capture the domain-specific knowledge. Furthermore, we jointly consider intent label and slot label sequence in the training process.
    \item Our method achieves state-of-the-art performance. Furthermore, we validate the effectiveness of the decoupling by extensive experiments.
\end{enumerate}

\section{Method}
In this section, we introduce the proposed Joint Modeling with Relationship Masking (JMRM) for few-shot intent classification and slot filling.

\subsection{Background}
\subsubsection{Problem Deﬁnition}
Natural language understanding (NLU) contains two related tasks: intent classification, which is a sentence-level text classification task, and slot filling, which is a word-level sequence labeling task.
Formally, a labeled sample can be represented as $(\boldsymbol{x},y,\boldsymbol{t})$,
where $ \boldsymbol{x} = \langle x_{1}, \cdots, x_{m} \rangle $ is an utterance with $m$ words, 
and $ y $ is the intent label of $ \boldsymbol{x} $,
and $ \boldsymbol{t} = \langle t_{1}, \cdots, t_{m} \rangle $ is the slot label sequence of $\boldsymbol{x}$.
Here, $ t_{i} $ is the slot label of $ x_{i} $.
A few-shot NLU model aims to predict both $y$ and $ \boldsymbol{t} $ for each $ \boldsymbol{x} $ according to the scarce labeled data in unseen target domains, with the experience transferred from source domains.


To match the source-domain setting with the target-domain setting, we construct multiple episodes in source domains.
Each episode contains a support set $\mathcal{S}$ with few labeled samples and a query set $\mathcal{Q}$ with samples to be predicted.
Formally, the support set is denoted as $ \mathcal{S} = \{(\boldsymbol{x}^{(n)}, y^{(n)}, \boldsymbol{t}^{(n)})\}^{|\mathcal{S}|}_{n=1} $,
and the query set is denoted as $ \mathcal{Q} = \{(\boldsymbol{x}^{(n)}, y^{(n)}, \boldsymbol{t}^{(n)})\}^{|\mathcal{Q}|}_{n=1} $,
where $n$ denotes the index of the sample.
During training, a model learns to predict labels of samples in $\mathcal{Q}$ based on $\mathcal{S}$.

\subsubsection{Preliminaries}
Before utilizing the two proposed modules, we obtain emission scores which are used to predict labels, based on Prototypical Networks~\cite{DBLP:conf/nips/SnellSZ17}.
Firstly, we calculate the label representation as to the mean embedding of support samples belonging to the same class:
\begin{equation} \label{cl}
C_{l} = \frac{1}{|\mathcal{S}_{l}|} \sum_{(\boldsymbol{x}, y, \boldsymbol{t}) \in \mathcal{S}_{l}} E(\boldsymbol{x}) ,
\end{equation}
\begin{equation}
C_{o} = \frac{1}{|\mathcal{S}_{o}|} \sum_{(\boldsymbol{x}, y, t_{i}) \in \mathcal{S}_{o}} E(x_{i}) ,
\end{equation}
where $E(\cdot)$ is a BERT~\cite{devlin-etal-2019-bert} based encoding function, and $ \mathcal{S}_{l} = \{(\boldsymbol{x}, y, \boldsymbol{t}) | y=l\} $ is the set of the samples with intent label $l$ in $\mathcal{S}$, and $ \mathcal{S}_{o} = \{(\boldsymbol{x}, y, t_{i}) | t_{i}=o\} $ is the set of the samples with slot label $o$ in $\mathcal{S}$.
Therefore, $C_{l}$ denotes the label representation of intent label $l$, and $C_{o}$ denotes the label representation of slot label $o$.

Then, for each query sample, we obtain the emission scores by calculating the similarity scores of the label representation and the sample embedding.
Specifically, we calculate the intent emission score $f_{l} \in \mathbb{R}^{1 \times Y}$ and the original slot emission score $f_{o} \in \mathbb{R}^{m \times T}$ as follows:
\begin{equation}
f_{l}(y=l) = \textrm{SIM}(E(\boldsymbol{x}), C_{l}),
\end{equation}
\begin{equation}
f_{o}(t_{i}=o) = \textrm{SIM}(E(x_{i}), C_{o}),
\end{equation}
where $Y$ and $T$ denote the number of intent and slot classes, respectively, and $\textrm{SIM}(\cdot, \cdot)$ is a similarity function.

\begin{figure*}[t]
\centering
\includegraphics[width=1.0\linewidth]{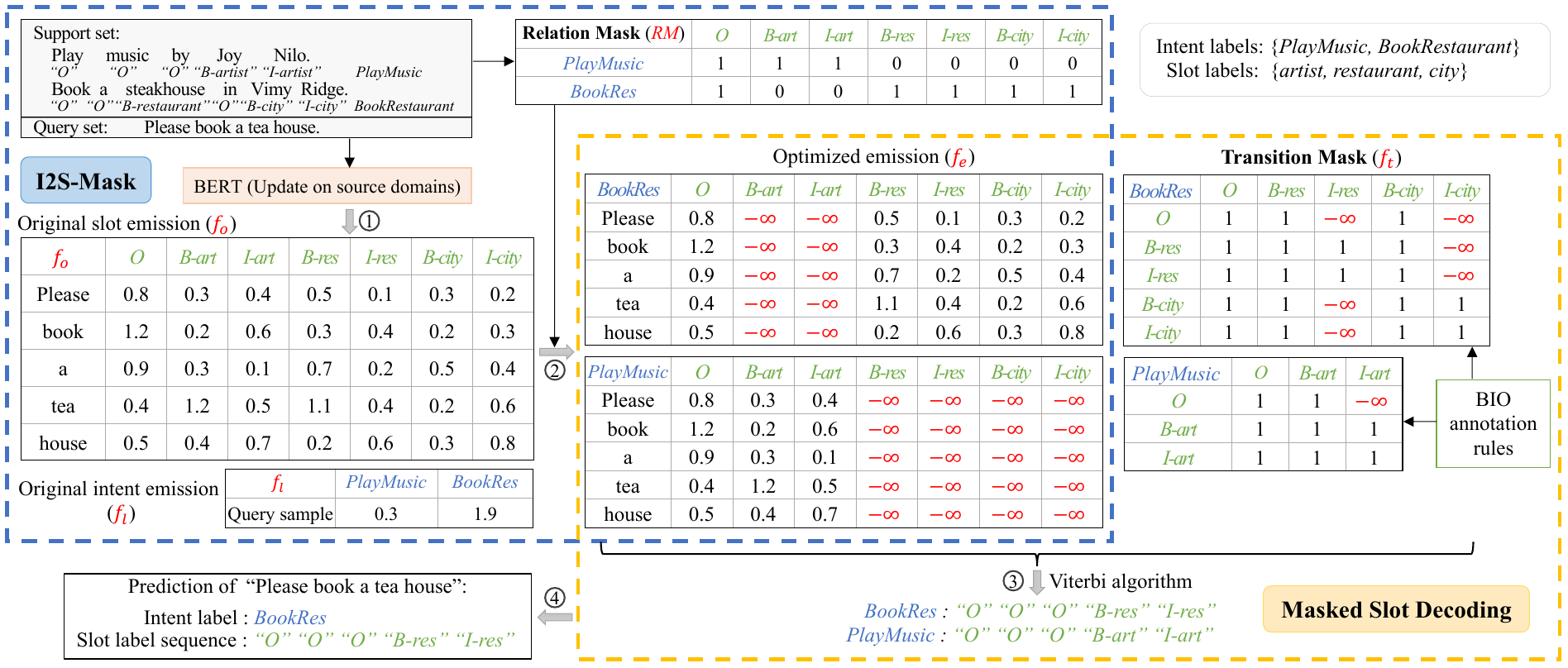}
\caption{Illustration of two main components of our method: the I2S-Mask module, which captures the domain-specific knowledge of intent-slot relation, and the Masked Slot Decoding module, which captures the domain-specific knowledge of slot-slot relation. Specifically, $Bookres$ denotes $BookRestaurant$, $art$ denotes $artist$, and $res$ denotes $restaurant$. We apply the two modules during both the training on source domains and the evaluation on target domains.}
\label{fig_method}
\end{figure*}

\subsection{I2S-Mask}
To capture the domain-specific knowledge of intent-slot relation, we propose I2S-Mask module.
It utilizes an intent-slot correlation score matrix as the domain-specific information of intent-slot relation, which regularizes the predicted intent and slot labels are related.

Specifically, in this paper, we define that an intent label and a slot label are \textbf{related} if they appear in a support sample simultaneously, such as $PlayMusic$ and $B\mbox{-}artist$ in the support set of Figure~\ref{fig_method}.
Otherwise, they are \textbf{unrelated}, such as $PlayMusic$ and $B\mbox{-}city$.

Firstly, we obtain an intent-slot relation mask, which is the intent-slot correlation score matrix summarized from the support set.
Formally, we denote the relation mask as RM $\in \mathbb{R}^{Y \times T}$.
It is shown in the I2S-Mask module of Figure~\ref{fig_method}, where 1 denotes the intent label in this row and the slot label in this column is related, and 0 denotes they are unrelated.

Then, we set the slot emission scores that are unrelated to the currently calculated intent label $l$ to negative infinity according to the RM:
\begin{equation}
f_{e}(t_{i}=o | y=l) = \begin{cases}
    f_{o}(t_{i}=o), \quad &\text{RM}_{l,o} = 1 \\
    -\infty, \quad &\text{otherwise}
\end{cases}
\end{equation}
where $f_{e} \in \mathbb{R}^{m \times T}$ and $\text{RM}_{l,o} = 1$ means that the intent label $l$ and slot label $o$ are related.

In summary, I2S-Mask calibrates the original slot emission ($f_{o}$) to the optimized emission ($f_{e}$) by the RM, which regularizes the predicted intent and slot labels are related.
Thus, during source-domain training, the model could decouple the source-domain-specific knowledge of intent-slot relation and the general semantic representation.
And during target-domain evaluation, without the source-domain-specific knowledge of intent-slot relation, the general semantic representation and the target-domain-specific knowledge of intent-slot relation make predictions more accurate.

\subsection{Masked Slot Decoding}
To capture the domain-specific knowledge of slot-slot relation, we propose Masked Slot Decoding module.
It utilizes a slot-slot constraint score matrix as the domain-specific information of slot-slot relation, which constrains the predicted slot label sequence is rational.

Specifically, a \textbf{rational} slot label sequence follows BIO annotation rules.
Formally, the $B\mbox{-}$ should be the label for the first word of a slot phrase, and the $I\mbox{-}$ should be the label for the other words of the slot phrase.
$O$ denotes $other$, indicating that the word is not important in the utterance.
For example, slot labels of ``Vimy Ridge" are ``$B\mbox{-}city \  I\mbox{-}city$".
And $I\mbox{-}city$ is allowed to follow $B\mbox{-}city$, but not $O$.

Firstly, we obtain a slot-slot transition mask, which is the slot-slot constraint score matrix summarized according to the BIO annotation rules of the slot labels in the support set.
Formally, we denote the transition mask as $f_{t} \in \mathbb{R}^{T \times T}$:
\begin{equation}
f_{t}(t_{i}=o_{2} | t_{i-1}=o_{1}) = \begin{cases}
    1, \quad &\text{BIO}_{o_{1},o_{2}} = 1 \\
    -\infty, \quad &\text{otherwise}
\end{cases}
\end{equation}
where $\text{BIO}_{o_{1},o_{2}} = 1$ means that slot label $o_{2}$ is allowed to follow slot label $o_{1}$.

Then, we utilize Viterbi algorithm~\cite{DBLP:conf/icml/LaffertyMP01} to predict the slot label sequence.
The input of Viterbi contains the optimized emission ($f_e$) and the unlearned transition mask ($f_t$).
The Viterbi algorithm utilizes the idea of dynamic programming to calculate the optimal slot label sequence among all possible predictions.

In summary, Masked Slot Decoding obtains the transition mask ($f_t$) according to BIO annotation rules, which regularizes the predicted slot label sequence is rational.
Thus, during source-domain training, the model could decouple the source-domain-specific knowledge of the slot-slot relation and the general semantic representation.
And during target-domain evaluation, without the source-domain-specific knowledge of slot-slot relation, the general semantic representation and the target-domain-specific knowledge of slot-slot relation make predictions more accurate.

\subsection{Joint Modeling}
\label{Joint Modeling}
Since the intent classification task and the slot filling
task are strongly related, it is beneficial to learn them jointly~\cite{weld2022survey}.
To this end, we jointly consider intent label and slot label sequence in the training process.

In the training process on source domains, firstly, given a query utterance $\boldsymbol{x}^{*}$ and the support set $\mathcal{S}$, we calculate the score of the intent label of $\boldsymbol{x}^{*}$ is $y$ and the slot label sequence of $\boldsymbol{x}^{*}$ is $\boldsymbol{t}$ as follows, with the source-domain-specific knowledge:
\begin{equation} \label{R}
\begin{split}
\begin{aligned}
R(y, \boldsymbol{t} | \boldsymbol{x}^{*}, \mathcal{S}) = \lambda \cdot f_{l}(y) + \sum_{i=1}^{m} (f_{e}(t_{i}|y) + f_{t}(t_{i}|t_{i-1})) ,
\end{aligned}
\end{split}
\end{equation}
where $\lambda$ is the weight of the intent score.

Secondly, we obtain the probability of the correct intent label $y^{*}$ and slot label sequence $\boldsymbol{t}^{*}$:
\begin{equation} \label{p}
p(y^{*}, \boldsymbol{t}^{*} | \boldsymbol{x}^{*}, \mathcal{S}) = \frac{\exp({R(y^{*}, \boldsymbol{t}^{*} | \boldsymbol{x}^{*}, \mathcal{S})})}{\sum_{y, \boldsymbol{t}}{\exp({R(y, \boldsymbol{t} | \boldsymbol{x}^{*}, \mathcal{S})})}}.
\end{equation}

Thirdly, we utilize a single cross-entropy loss function to update the few-shot NLU model:
\begin{equation}
\mathcal{L}  = -\log p(y^{*}, \boldsymbol{t}^{*} | \boldsymbol{x}^{*}, \mathcal{S}).
\end{equation}

During the evaluation on target domains, the transferred model predicts the intent label and slot label sequence simultaneously as follows, with the target-domain-specific knowledge:
\begin{equation}
y^{*}, \boldsymbol{t}^{*} = \mathop{\arg\max}\limits_{y,\boldsymbol{t}} p(y, \boldsymbol{t} | \boldsymbol{x}^{*}, \mathcal{S}).
\end{equation}

The joint modeling considers both intent classification and slot filling tasks simultaneously.
And as validated in Figure~\ref{fig_analysis_jointbetter}, the two related tasks guide each other to obtain better performance.

\section{Experiments}

\subsection{Dataset and Domain Construction}
\subsubsection{Dataset.}
We conduct extensive experiments on two natural language understanding (NLU) benchmarks: the Snips dataset~\cite{Snips} and the FewJoint dataset~\cite{FewJoint}. 
Specifically, Snips is an English dataset with 7 intent classes and 48 slot classes. 
FewJoint is a Chinese dataset with 141 intent classes and 193 slot classes.

\begin{table*}[t]
    \centering
    \begin{tabular}{lcccccc}
        \toprule
        \multirow{2}*{\textbf{Models}} & \multicolumn{3}{c}{\textbf{Snips}} & \multicolumn{3}{c}{\textbf{FewJoint}} \\
        \cmidrule(lr){2-4} \cmidrule(lr){5-7} & Intent Acc. & Slot F1 & Joint Acc. & Intent Acc. & Slot F1 & Joint Acc. \\
        \midrule
        SepProto & \textbf{98.23$_{\pm0.66}$} & 43.90$_{\pm1.98}$ & \ \ 9.47$_{\pm2.10}$ & \textbf{66.35}$_{\pm0.51}$ & 27.24$_{\pm1.10}$ & 10.92$_{\pm0.89}$ \\
        JointProto & 92.57$_{\pm0.57}$ & 42.63$_{\pm2.03}$ & \ \ 7.35$_{\pm1.70}$ & 58.52$_{\pm0.28}$ & 29.49$_{\pm1.01}$ & \ \ 9.64$_{\pm0.47}$\\
        ConProm & 96.67$_{\pm1.45}$ & 53.05$_{\pm0.81}$ & 21.72$_{\pm0.97}$ & 65.26$_{\pm0.23}$ & 33.09$_{\pm1.66}$ & 16.32$_{\pm0.75}$ \\
        ConProm+FT+TR & 90.45$_{\pm0.52}$ & 56.04$_{\pm1.75}$ & 27.80$_{\pm2.33}$ & 63.67$_{\pm0.94}$ & 42.44$_{\pm0.51}$ & 27.72$_{\pm0.95}$  \\
        \textbf{JMRM}(Ours) & 93.71$_{\pm0.74}$ & \textbf{66.55}$_{\pm0.44}$ & \textbf{41.76}$_{\pm1.25}$ & 65.97$_{\pm0.70}$ & \textbf{62.24}$_{\pm1.17}$ & \textbf{42.20}$_{\pm0.88}$ \\
        \bottomrule
    \end{tabular}
    \caption{Results of 1-shot natural language understanding task on Snips and FewJoint. TR denotes the BIO transition rule of adjacent slots, e.g. slot label ``I'' is not allowed to follow ``O''. FT denotes fine-tuning model on target domains.}
    \label{tab:1-shot}
\end{table*}

\begin{table*}[t]
    \centering
    \begin{tabular}{lcccccc}
        \toprule
        \multirow{2}*{\textbf{Models}} & \multicolumn{3}{c}{\textbf{Snips}} & \multicolumn{3}{c}{\textbf{FewJoint}} \\
        \cmidrule(lr){2-4} \cmidrule(lr){5-7} & Intent Acc. & Slot F1 & Joint Acc. & Intent Acc. & Slot F1 & Joint Acc. \\
        \midrule
        SepProto & \textbf{99.53$_{\pm0.11}$} & 53.28$_{\pm1.85}$ & 14.40$_{\pm3.00}$ & 75.64$_{\pm1.51}$ & 36.08$_{\pm0.65}$ & 15.93$_{\pm1.85}$ \\
        JointProto & 99.17$_{\pm0.09}$ & 50.63$_{\pm2.01}$ & 13.40$_{\pm1.44}$ & 70.93$_{\pm2.45}$ & 39.47$_{\pm1.05}$ & 14.48$_{\pm1.11}$ \\
        ConProm & 98.50$_{\pm0.42}$ & 61.03$_{\pm1.77}$ & 32.20$_{\pm2.06}$  & 78.05$_{\pm1.04}$ & 39.40$_{\pm1.75}$ & 24.18$_{\pm1.29}$ \\
        ConProm+FT+TR & 98.40$_{\pm0.20}$ & 72.98$_{\pm0.41}$ & 52.95$_{\pm0.85}$ & 78.43$_{\pm1.86}$ & 69.44$_{\pm0.39}$ & 46.54$_{\pm0.72}$ \\
        \textbf{JMRM}(Ours) & 97.84$_{\pm0.69}$ & \textbf{77.82}$_{\pm1.47}$ & \textbf{59.24}$_{\pm2.25}$ & \textbf{82.26}$_{\pm0.64}$ & \textbf{76.15}$_{\pm0.58}$ & \textbf{60.79}$_{\pm0.30}$ \\
        \bottomrule
    \end{tabular}
    \caption{Results of 5-shot natural language understanding task on Snips and FewJoint.}
    \label{tab:5-shot}
\end{table*}

\subsubsection{Domain construction.}
We aim to provide a comprehensive description of domain construction across three levels of granularity: domain types, the quantity of episodes within each domain type, and episode construction.

In our experiments, we construct the training domains as source domains to update model parameters, the developing domains to select the best model, the testing domains as target domains to evaluate the performance.
For Snips, we construct a training domain with 3 intent classes PlayMusic, AddToPlayList, BookRestaurant, a developing domain with 2 intent classes RateBook, SearchScreeningEvent, and a testing domain with 2 intent classes GetWeather, SearchCreativeWork.
For FewJoint, we construct 38 training domains, 5 developing domains, and 9 testing domains.
Here, each domain has multiple intent labels and slot labels. 

For the quantity of episodes in each domain, on Snips, we construct 200, 50, 10 episodes for the training, developing and testing domains, respectively.
On FewJoint, we use the original few-shot episodes.

Formally, each episode contains a support set with few labeled samples and a query set with some samples to be predicted.
For constructing episodes, we adopt the Mini-Including Algorithm~\cite{hou-etal-2020-shot} that constructs a support set in the $K$-shot setting following two criteria:
(1) Each slot class contains at least $K$ examples.
(2) Removing any utterance would cause the former be not hold.

\subsection{Baselines and Evaluation Metrics}
We compare our method with competitive few-shot NLU baselines:
\begin{itemize}
    \item \textbf{SepProto} utilizes Prototypical Networks~\cite{DBLP:conf/nips/SnellSZ17} for the intent classification and slot filling tasks with the separate BERT~\cite{devlin-etal-2019-bert} embedding. The model is trained on the training domains and then evaluated directly on the unseen testing domains without fine-tuning.
    \item \textbf{JointProto}~\cite{krone-etal-2020-learning} jointly learns the intent and slot representations by sharing a single BERT encoder on source domains, without fine-tuning on target domains.
    \item \textbf{ConProm}~\cite{hou-etal-2021-learning} merges the intent and slot representations into one space and learns the representations by contrastive learning.
    \item \textbf{ConProm+TR+FT}, where FT denotes fine-tuning models, and TR denotes the transition rules of BIO annotation, which ban illegal slot predictions from left to right during target-domain evaluation.
\end{itemize}

There are 3 evaluation metrics in the NLU task: intent accuracy (Intent Acc), slot F1-score (Slot F1), and joint accuracy (Joint Acc). 
The Joint Acc means that both the predicted intent label and slot label sequence of a query utterance are correct, which is the most important metric.
In our experiments, we select the model on the developing domains according to Joint Acc, and then report the performance of the model on testing domains as final results. 
To relieve the non-deterministic model training~\cite{reimers-gurevych-2017-reporting}, we report the average score of 5 experiments with different random seeds for every setting.

\begin{table*}[t]
    \centering
    \begin{tabular}{lccllllll}
        \toprule
        \multirow{2}*{\textbf{Models}} & \multicolumn{2}{c}{\textbf{Training}} & \multicolumn{3}{c}{\textbf{1-shot}} & \multicolumn{3}{c}{\textbf{5-shot}} \\
        \cmidrule(lr){2-3} \cmidrule(lr){4-6} \cmidrule(lr){7-9} & +I2S & +MSD & Cos & L2 & VPB & Cos & L2 & VPB \\
        \midrule
        (JointProto) & & & 27.54 & 26.11 & 26.92 & 39.70 & 37.69 & 43.02 \\
        (JointProtoI2S) & $\surd$ & & 33.30$_{\uparrow 5.76}$ & 33.46$_{\uparrow 7.35}$ & 28.89$_{\uparrow 1.97}$ & 48.37$_{\uparrow 8.67}$ & 50.93{$_{\uparrow 13.24}$} & 52.25$_{\uparrow 9.23}$ \\
        \midrule
        (\textbf{JM}) & & & 24.29 & 25.79 & 24.98 & 38.93 & 37.24 & 45.37 \\
        (\textbf{JMI2S}) & $\surd$ & & 36.56$_{\uparrow 12.27}$ & 37.09$_{\uparrow 11.30}$ & 37.78$_{\uparrow 12.80}$ & 49.31$_{\uparrow 10.38}$ & 50.86$_{\uparrow 13.62}$ & 54.57$_{\uparrow 9.20}$ \\
        (\textbf{JMMSD}) & & $\surd$ & 32.99$_{\uparrow 8.70}$ & 32.11$_{\uparrow 6.32}$ & 32.90$_{\uparrow 7.92}$ & 47.34$_{\uparrow 8.41}$ & 46.54$_{\uparrow 9.30}$ & 46.42$_{\uparrow 1.05}$ \\
        \textbf{JMRM} & $\surd$ & $\surd$ & 41.35$_{\uparrow 17.06}$ & 41.75$_{\uparrow 15.96}$ & 41.97$_{\uparrow 16.99}$ & 59.54$_{\uparrow 20.61}$ & 58.82$_{\uparrow 21.58}$ & 60.79$_{\uparrow 15.42}$ \\
        \bottomrule
    \end{tabular}
    \caption{Ablation study of the I2S-Mask (I2S) module and the Masked Slot Decoding (MSD) module on FewJoint, with three different similarity functions, including Cos, L2 and VPB. JM denotes joint modeling.}
    \label{tab:ablation}
\end{table*}

\subsection{Implementation Details}
For a fair comparison, the hyperparameters in our experiments are set the same as in baselines. 
The batch size is 4 and the learning rate is $10^{-5}$. 
We set a single BERT~\cite{devlin-etal-2019-bert} as the embedding function of intents and slots, where the intent representation of an utterance is the average of all word embedding in the utterance.
The weight of the intent score $\lambda$ in Eq.~\ref{R} is 1.
We use ADAM~\cite{DBLP:journals/corr/KingmaB14} to update model parameters and transfer the learned general semantic representation from source domains to target domains without fine-tuning. 
To observe the robustness of our method to the similarity function, we utilize 3 different similarity functions in our experiments, including cosine (Cos), euclidean distance (L2), and the vector projection with bias (VPB)~\cite{zhu2020vector}.

We conduct experiments of 1-shot on GeForce GTX 1080, and 5-shot on GeForce RTX 3090.
Training on Snips takes 2.8 hours on average.
Due to early convergence, on FewJoint, the 1-shot training takes 0.8 hours and the 5-shot training takes 2.2 hours.

\begin{table}[t]
    \centering
    \resizebox{0.85\linewidth}{!}{
    \begin{tabular}{lcccc}
        \toprule
        \multirow{2}*{\textbf{Models}} & \multicolumn{2}{c}{\textbf{Snips}} & \multicolumn{2}{c}{\textbf{FewJoint}} \\
        \cmidrule(lr){2-3} \cmidrule(lr){4-5} & 1-shot & 5-shot & 1-shot & 5-shot \\
        \midrule
        JointProto & \ \ 7.35 & 13.40 & \ \ 9.64 & 14.48 \\
        +RM & 21.32 & 19.13 & 30.11 & 40.18 \\
        \midrule
        ConProm & 21.72 & 32.20 & 16.32 & 24.18 \\
        +RM & 30.65 & 47.30 & 34.21 & 37.72 \\
        \midrule
        JM & 30.42 & \ 38.37 & 24.29 & 38.93 \\
        +RM & 38.40 & \ 51.90 & 36.71 & 53.53 \\
        \bottomrule
    \end{tabular}
    }
    \caption{Exploration of the effect of the target-domain-specific knowledge, with Joint Acc metrics.}
    \label{tab:RMprediction}
\end{table}

\subsection{Results and Analysis}
To verify the proposed method, we conduct extensive experiments on the NLU task, in 1-shot and 5-shot settings. 
Moreover, we present some auxiliary experiments to analyze the effectiveness of the proposed decoupling and joint modeling for few-shot NLU.

\subsubsection{Result of 1-shot}
The experimental results of the NLU task in the 1-shot setting are shown in Table~\ref{tab:1-shot}. 
For Joint Acc, the most important metric, our method achieves state-of-the-art performance.
Specifically, on Snips we improve Joint Acc from 27.80\% to 41.76\%, and on FewJoint we improve Joint Acc from 27.72\% to 42.20\%.

Interestingly, the Intent Acc results of SepProto are highest on both Snips and FewJoint.
This is because SepProto has two BERT models, which learn intent embedding and slot embedding separately.
In contrast, our method is not competitive on Intent Acc.
This is because with sharing a single BERT encoder when we focus more on the slot filling task in the 1-shot setting, the performance of intent classification tends to decrease. 
The same phenomenon has also been observed in other works~\cite{hou-etal-2021-learning, krone-etal-2020-learning}.

\subsubsection{Result of 5-shot}
The experimental results of the NLU task in the 5-shot setting are shown in Table~\ref{tab:5-shot}. 
On the metric of Joint Acc, our method also achieves state-of-the-art performance.
Specifically, on Snips we improve Joint Acc from 52.95\% to 59.24\%, and on FewJoint we improve Joint Acc from 46.54\% to 60.79\%.

Notably, our method achieves the best Intent Acc on FewJoint.
This profits from the joint modeling, which makes the two related tasks guide each other.
As an explanation, we explicitly add the intent guidance for slots and slot guidance for intents into training objectives.
Here, an improvement in the performance of slot filling beyond a certain threshold can facilitate the learning process of intent classification.

\subsubsection{Ablation study}
To analyze contributes of I2S-Mask (I2S) and Masked Slot Decoding (MSD), we conduct ablation experiments.
Table~\ref{tab:ablation} shows the Joint Acc results of 1-shot and 5-shot on FewJoint.
JM is the proposed joint modeling, without I2S and MSD.
+I2S denotes introducing the intent-slot relation mask in I2S into training objectives, and +MSD denotes introducing the slot-slot transition mask in MSD into training objectives.
Furthermore, to observe the robustness of our method to the similarity function, we utilize three similarity functions, including Cos, L2, and VPB.

Remarkably, I2S and MSD improve performance in all settings.
Specifically, compared with JointProto, JointProtoI2S performs better on all similarity functions, and the highest improvement of 1-shot is 7.35 points and that of 5-shot is 13.24 points, both on the L2 similarity function.
And compared with JM, JMI2S and JMMSD perform better on all similarity functions.
Interestingly, JMI2S is better than JMMSD.
This is because, in I2S, the intents and slots calculated in Eq.~\ref{R} are all related, making the bi-directional guidance between intents and slots more helpful. 
And compared with JM, JMRM improves the performance by up to 17.06 points with Cos in the 1-shot setting and 21.58 points with L2 in the 5-shot setting.

Furthermore, in most instances, VPB is better than Cos and L2.
The explanation is as follows.
VPB is a metric function proposed by~\citeauthor{zhu2020vector}, and its formula is: 
$\textrm{SIM}(\boldsymbol{e}, C) = \boldsymbol{e}^{\top} \frac{C}{||C||} - \frac{1}{2} ||C||$
, where $\boldsymbol{e}$ denotes the sample embedding and $C$ denotes the label representation. 
The regularization of C in VPB prevents excessive influence when it becomes too large and reduces false positive errors, thereby improving performance~\cite{zhu2020vector}. 

\subsubsection{Analysis}
\begin{figure}[t]
\centering
\includegraphics[width=0.9\columnwidth]{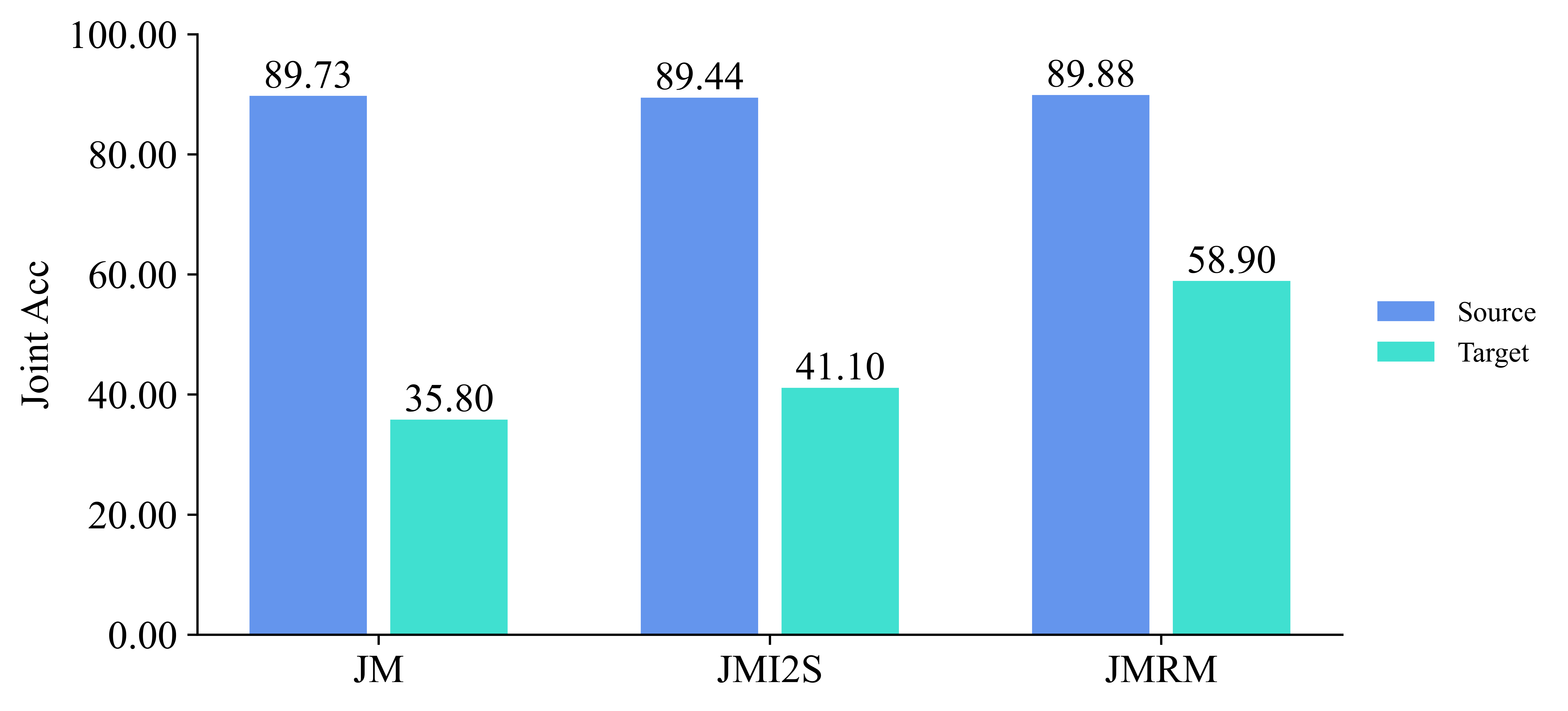}
\caption{Performance of different methods on source and target domains, in 5-shot setting on Snips.}
\label{fig_decoupling}
\end{figure}

\begin{figure}[t]
\centering
\includegraphics[width=0.9\columnwidth]{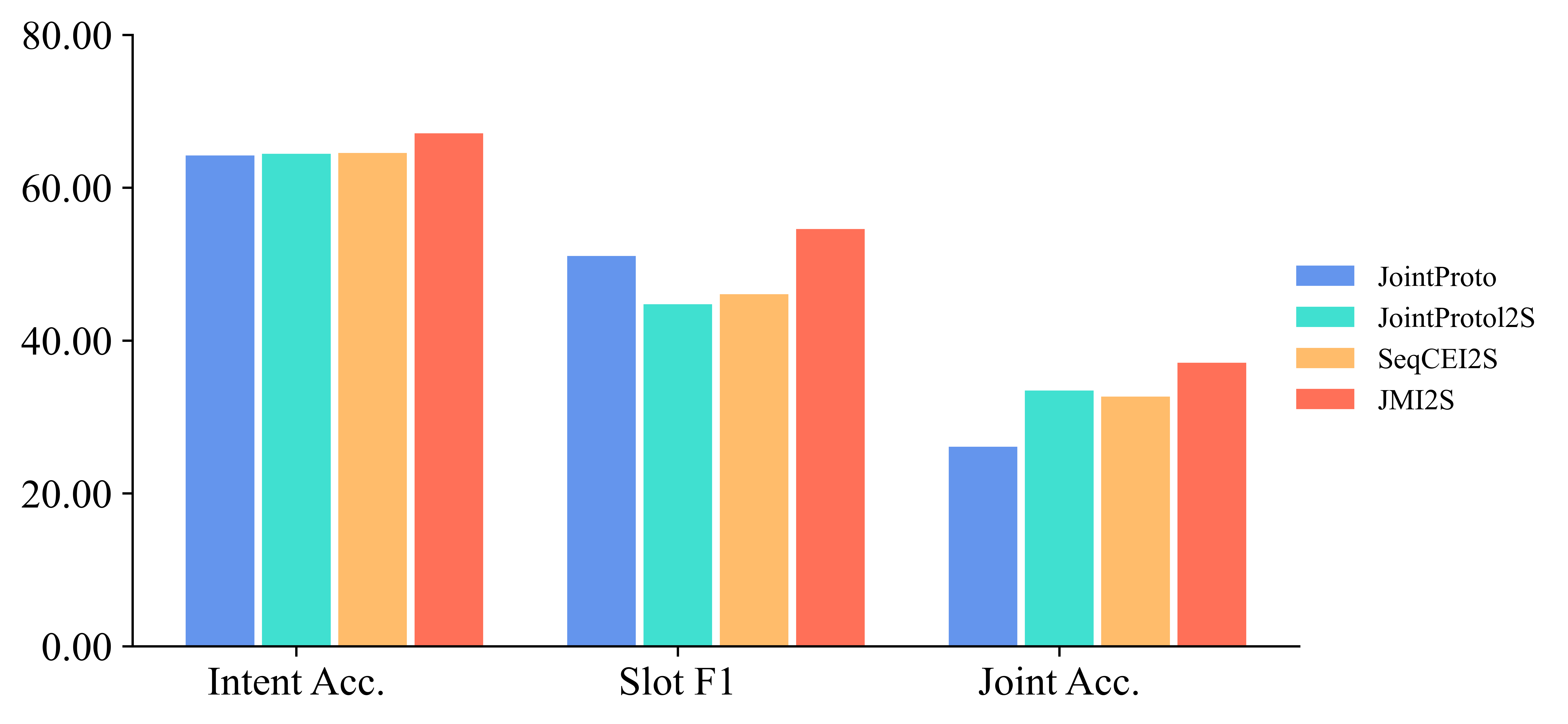}
\caption{Analysis of the benefits of joint modeling.}
\label{fig_analysis_jointbetter}
\end{figure}

\begin{itemize}
    \item \textbf{Domain-specific knowledge helps predictions be more accurate.} To explore the effect of using the target-domain-specific knowledge, we conduct the experiments in Table~\ref{tab:RMprediction}. 
+RM denotes that a model utilizes I2S and MSD only during target-domain evaluation.

Remarkably, +RM improves the performance of all models.
For example, in the 5-shot setting on FewJoint, +RM helps JointProto, ConProm, and JM improve the Joint Acc metrics up to 40.18\%, 37.72\%, and 53.53\%, respectively.
This is because +RM captures the target-domain-specific knowledge, which makes predictions satisfy the relationship constraints.
Thus, domain-specific knowledge makes predictions more accurate.

Moreover, our method JM with RM achieves the best results in all settings compared to other methods with RM.
This indicates that the proposed joint modeling is efficient for few-shot NLU.
Furthermore, the JMRM is better than JM+RM, which shows that the decoupling during source-domain training is efficient.

    \item \textbf{Decoupling makes transferring more efficient.}
To investigate the effectiveness of decoupling the general semantic representation and the domain-specific knowledge, we conduct the experiment in Figure~\ref{fig_decoupling}.
Here, we list Joint Acc results of three different training methods, in the 5-shot setting on Snips.
JMI2S uses only the I2S module during training, decoupling the domain-specific knowledge of intent-slot relation.
JMRM uses both the I2S module and MSD module during training, decoupling the domain-specific knowledge of both intent-slot relation and slot-slot relation.

Notably, JM, JMI2S, and JMRM all perform similarly on source domains, which indicates that these three methods have learned source-domain experience of roughly equivalent magnitudes.
During target-domain evaluation, JM transfers all source-domain experience as a whole, and JMI2S and JMRM transfer the decoupled general semantic representation.
Since JMI2S is better than JM and JMRM is better than JMI2S, we conclude that, due to the gaps among different domains, the source-domain-specific knowledge will suffer the model’s performance on target domains.
Thus, decoupling the general semantic representation and the domain-specific knowledge makes transferring more efficient.

\item \textbf{JM achieves bi-directional guidance.}
To explore the effect of the proposed joint modeling, we analyze the experiments in Figure~\ref{fig_analysis_jointbetter}.
JointProtoI2S first predicts intent labels, and then predicts slot labels according to the relation mask.
This means that JointProtoI2S uses intent information to guide the slot prediction.
Significantly, JointProto compares with JointProtoI2S, the Intent Acc is about the same, the Slot F1 is a little better, but the Joint Acc is lower.
It means that in JointProto, there are some samples whose slot predictions are correct, but their intent labels are wrong. 
Therefore, we conjecture that slot information can also guide the learning of intents.

To verify our conjecture, we further design SeqCEI2S which is different from JMI2S only in the joint modeling. 
Specifically, the loss of JointProtoI2S is:
$ \mathcal{L} = \mathcal{L}_{intent} + \sum_{i}{\mathcal{L}_{slot_i}} $, 
and the loss of SeqCEI2S is: 
$ \mathcal{L} = \mathcal{L}_{intent} + \mathcal{L}_{slot\_seq} $, 
and the loss of JMI2S is:
$ \mathcal{L} = \mathcal{L}_{intent + slot\_seq} $,
where $slot\_seq$ denotes predicting the slot label sequence using the Viterbi algorithm.
The results show that the Intent Acc of JMI2S is higher than SeqCEI2S, which indicates that the intent task is indeed influenced by the guidance from the slot task. 
JMI2S also has higher Slot F1 and Joint Acc than others.
Thus we conclude that joint modeling achieves bi-directional guidance of both intents and slots, resulting in better performance.

\end{itemize}

\section{Related work}
\subsection{Natural language understanding}
Natural language understanding (NLU) is an important component of dialogue systems, including the intent classification task and the slot filling task.
There are a lot of NLU methods~\cite{goo-etal-2018-slot, wang-etal-2018-bi, liu-etal-2020-coach, ma-etal-2021-intention, rosenbaum-etal-2022-linguist, zheng2023hit, ma-etal-2022-unitranser}.
\citeauthor{qin-etal-2019-stack} directly took the output of the intent task as the input to the slot task.
\citeauthor{chen2019bert} utilized a shared BERT encoder to jointly learn intents and slots.
\citeauthor{conf/icassp/QinLCKZ021} utilized multiple label attention layers and co-interactive attention layers to jointly encode the intent and slot representations.
\citeauthor{DBLP:conf/ijcai/QinXC021} listed some new areas and challenges related to NLU.
However, these methods are limited in performance in specialist areas where data are highly variable and samples are difficult to collect.

\subsection{Few-shot learning}
Few-shot learning aims to learn models based on a few samples.
Generally, the models are first trained on sufficient source domains, then evaluated on unseen target domains with few labeled data. 
Few-shot learning approaches in natural language processing mainly include three approaches.
In the fine-tuning-based approaches~\cite{sun2019meta, shen2021partial}, MAML~\cite{finn2017model} trained model parameters such that a small number of gradient updates will lead to fast learning on a new task with few labeled data, and ULMFiT~\cite{howard-ruder-2018-universal} fine-tuned the language model and the classifier on target tasks.
In the prompting-based approaches~\cite{ma-etal-2021-frustratingly, gao-etal-2021-making, li-liang-2021-prefix}, PET~\cite{schick-schutze-2021-just} converted the text classification task into a masked language model task for few-shot learning.
In the metric learning-based approaches~\cite{DBLP:conf/nips/SnellSZ17,hou-etal-2020-shot,yuan2021dcen}, VPB~\cite{zhu2020vector} utilized the projections of contextual word representations on each normalized label representation as the word-label similarity.
However, these methods are not designed for intent and slot tasks in NLU. 
Our method specially solves the intent and slot problems by considering the relevance of these two tasks.

Recently, few-shot NLU attracts widespread attention due to the data scarcity problem~\cite{rongali-etal-2023-low, kwon2023zeroshot, gangadharaiah2022zeroshot}.
ZEROTOP~\cite{mekala-etal-2023-zerotop} utilized large language models to complete NLU tasks and designed different prompts for different intent/slot labels.
However, the large language model is worse at slot filling~\cite{he2023chatgpt, qin-etal-2023-chatgpt}.
Thus, it is reasonable to use classical few-shot learning methods for few-shot NLU.
\citeauthor{krone-etal-2020-learning} utilized classical few-shot learning methods MAML and Prototypical Networks to solve the few-shot intent classification and slot filling problem.
However, these methods did not take into account the relationship between the two tasks.
Therefore, few-shot joint learning of intents and slots becomes popular~\cite{basu2021semi}.
\citeauthor{hou-etal-2021-learning} merged the intent and slot representations into one space with the attention mechanism.
\citeauthor{liu-etal-2021-explicit-joint} utilized contrastive learning to learn the intent and slot representations jointly.
However, the transferred experience remains difficult due to the gaps between different domains.
Therefore, we consider decoupling the general semantic representation and the domain-specific knowledge to relieve the transfer disaster.

\section{Conclusion}
In this paper, we propose a new method, \textbf{JMRM}, for few-shot intent classification and slot filling.
It explicitly decouples general semantic representation and domain-specific knowledge, and only transfers the general semantic representation to target domains.
Specifically, to capture the domain-specific knowledge, we design the I2S-Mask and Masked Slot Decoding modules, which utilize two relationship score matrices to regularize predictions.
Experiments validate that decoupling makes transferring more efficient.

\section{Acknowledgments}
We are grateful to the anonymous reviewers for their valuable comments.
This work is supported by National Natural Science Foundation of China under grants 62376103, 62302184, 62206102, and Science and Technology Support Program of Hubei Province under grant 2022BAA046.

\bibliography{aaai24}

\end{document}